\documentclass{article}
\usepackage{spconf,amsmath,amssymb,graphicx}
\usepackage{array}
\newcolumntype{P}[1]{>{\centering\arraybackslash}p{#1}}

\title{Deep word embeddings for visual speech recognition}
\name{Themos Stafylakis and Georgios Tzimiropoulos}
\address{Computer Vision Laboratory, University of Nottingham, UK}
\begin{document}
%
\maketitle
\begin{abstract}
In this paper we present a deep learning architecture for extracting word embeddings for visual speech recognition. The embeddings summarize the information of the mouth region that is relevant to the problem of word recognition, while suppressing other types of variability such as speaker, pose and illumination. The system is comprised of a spatiotemporal convolutional layer, a Residual Network and bidirectional LSTMs and is trained on the Lipreading in-the-wild database. We first show that the proposed architecture goes beyond state-of-the-art on closed-set word identification, by attaining 11.92\% error rate on a vocabulary of 500 words. We then examine the capacity of the embeddings in modelling words unseen during training. We deploy Probabilistic Linear Discriminant Analysis (PLDA) to model the embeddings and perform low-shot learning experiments on words unseen during training. The experiments demonstrate that word-level visual speech recognition is feasible even in cases where the target words are not included in the training set.
\end{abstract}
\section{Introduction}
\label{sec:intro}
Automatic speech recognition (ASR) is witnessing a renaissance, which can largely be attributed to the advent of deep learning architectures. Methods such as Connectionist Temporal Classification (CTC) and attentional encoder-decoder facilitate ASR training by eliminating the need of frame-level senone labelling, \cite{Prabhavalkar2017} \cite{Takaaki2017} while novel approaches deploying words as recognition units are challenging the conventional wisdom of using senones as recognition units, \cite{AcWordEmb16} \cite{audhkhasi2017} \cite{soltau2016neural} \cite{bengio2014word}. 
In parallel, architectures and learning algorithms initially proposed for audio-based ASR are combined with powerful computer vision models and are finding their way to lipreading and audiovisual ASR, \cite{chung2017lipsent} \cite{chung2017profile} \cite{assael2016lipnet} \cite{Thangthai2017} \cite{Wand2017} \cite{Potamianos2017} \cite{petridisend}. 

Motivated by this recent direction in acoustic LVCSR of considering words as recognition units, we examine the capacity of deep architectures for lipreading in extracting word embeddings. Yet, we do not merely address word identification with large amounts of training instances per target word; we are also interested in assessing the {\em generalizability} of these embeddings to words unseen during training. This property is crucial, since collecting several hundreds of training instances for all the words in the dictionary is impossible. 

To this end, we train and test our architecture on the LipReading in-the-Wild database (LRW, \cite{chung2016lip}), which combines several desired properties, such as relatively high number of target words (500), high number or training examples per word (between 800 and 1000), high speaker and pose variability, non-laboratory recording conditions (excerpts from BBC-TV) and target words that are part of segments of continuous speech of fixed 1.16s duration. We examine two settings; standard closed-set word {\it identification} using the full set of training instances per target word, and {\em low-shot learning} where the training and test words come from disjoint sets. For the latter setting, a PLDA model is used on the embedding domain that enables us to estimate class (i.e. word) conditional densities and evaluate likelihood ratios. Our proposed architecture is an improvement of the one we recently introduced in \cite{Stafy2017} which obtains state-of-the-art results on LRW even without the use of word boundaries.

The rest of the paper is organized as follows. In Sect. \ref{sect:Proposed} we provide a detailed description of the architecture, together with information about the training strategy and the use of word boundaries. In Sect. \ref{Sub:ID} we show results on word identification obtained when the model is training on all available instances, while in Sect. \ref{Sub:LS} we present results on two low-shot learning experiments. Finally, conclusion and directions for future work are given in Sect. \ref{sect:Conc}.    
\section{PROPOSED NETWORK ARCHITECTURE}
\label{sect:Proposed}
In this section we describe the network we propose, together with details regarding the preprocessing, the training strategy and loss function.
\subsection{Detailed description of the network}
\begin{figure}[!htbp]
\centering
\includegraphics[width=2.90in]{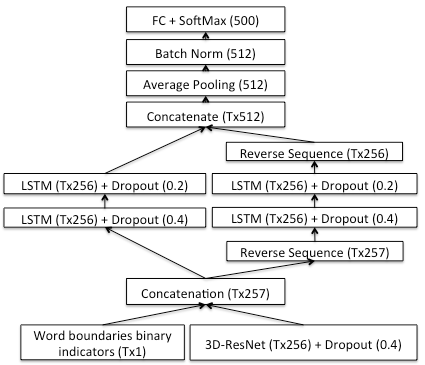}
\vspace{-2mm}\caption{The block-diagram of the proposed network.}
\label{diagram}
\end{figure}

The proposed architecture is depicted in Fig. \ref{diagram} and it is an extension of the one we introduced in \cite{Stafy2017}. The main differences are (a) the use of a smaller ResNet that (18 rather than 34-layer) that reduces the number of parameters from $\sim 24$M to $\sim 17$M, (a) the use of a pooling layer for aggregating information across time steps extracting a single embedding per video, (b) the use of dropouts and batch normalization at the back-end, and (c) the use of word boundaries which we pass to the backend as additional feature.
\subsubsection{ResNet with spatiotemporal front-end} The frames are passed through a Residual CNN (ResNet), which is a $18$-layer convolutional network with skip connections and outputs a single $256$-sized per time step, i.e. a $T\times256$ tensor ($T=29$ in LRW). There are two differences from the ImageNet 18-layer ResNet architecture, \cite{he2016identity}; (a) the first 2D convolutional layer has been replaced with a 3D (i.e. spatiotemporal) convolutional layer with kernel size $5\times7\times7$ (time$\times$width$\times$height) and the same holds for the first batch normalization and max pooling layers (without reducing the time resolution), and (b) the final average pooling layer (introduced for object recognition and detection) has been replaced with a fully connected layer, which is more adequate for face images that are centred. The model is trained from scratch, since pretrained models cannot be deployed due to the spatiotemporal front-end.
\subsubsection{Backend, embedding layer and word boundaries} The backend is composed of a two-layer BiLSTM followed by an {\em average pooling} layer that aggregates temporal information enabling us to extract a single $512$-size representation vector (i.e. embedding) for each video. The two-layer BiLSTM differs from the usual stack of two BiLSTM model; we obtained significantly better results by concatenating the two directional outputs only at the output of the second LSTMs. The backend receives as input the collection of $256$-size features extracted by the ResNet (CNN-features) concatenated with a binary variable indicating whether or not the frame lies inside or outside the {\em word boundaries}. We choose to pass the word boundaries as a feature because (a) dropping out potentially useful information carried in the out-of-boundaries frames is not in the spirit of deep learning, and (b) the gating mechanism of LSTMs is powerful enough to make use of it in an optimal way. 

Dropouts with $p=0.4$ are applied to the inputs of each LSTM (yet not to the recurrent layer, see \cite{LSTMdrop2017}), with the mask being fixed across features of the same sequence. Finally, batch normalization is applied to the embedding layer, together with a dropout layer with $p=0.2$, \cite{ioffe2015batch}.

\subsection{Preprocessing, loss and optimizer}
The preprocessing and data augmentation are identical to \cite{Stafy2017}. Moreover, as in \cite{Stafy2017}, we start training the network using a simpler convolutional backend which we subsequently replace with the LSTM backend, once the ResNet is properly initialized\footnote{Code and pretrained models in Torch7 are available at https://github.com/tstafylakis/Lipreading-ResNet}. Contrary to \cite{Stafy2017}, we use the Adam optimizer \cite{Adam}, with initial learning rate equal to $3\times10^{-3}$ and final equal to $10^{-5}$, and we drop it by a factor of 2 when no progress is attained for 3 consecutive epochs on the validation set. The algorithm typically converges after 50-60 epochs.  
We train the network using the cross-entropy criterion with softmax layer over training words. This criterion serves for both tasks we examine, i.e. closed-set word identification and low-shot learning, while its generalizability to unseen classes is in general equally good compared to other pairwise losses (e.g. contrastive loss), \cite{AcWordEmb16} \cite{DeepFace2014} \cite{Parkhi2015} \cite{Snyder2017}.

\section{EXPERIMENTAL RESULTS}
We demonstrate the effectiveness of the proposed architectures with respect to two experimental settings. The first is the standard closed-set identification task in which the network is trained with all available instances per target word (between 800 and 1000, \cite{chung2016lip}). The second setting aims to address to problem of word recognition on words unseen during training. The few training instances of the target words are merely used for estimating word-conditional densities on the embedding domain, via PLDA. To this end, the network is trained using a subset of 350 words of LRW and the test pairs are drawn from the remaining 150 words. 
\subsection{Closed-set word identification}
\label{Sub:ID}
For our first experiments in word identification, the reduced word set (consisting of the 350 words out of 500) will be used for both training and testing. These networks will also be used for low-shot learning on the remaining 150 words. Our proposed networks is retrained and tested on the full 500-word set. 
\subsubsection{Baseline and state-of-the-art}
We compare our architecture with two approaches which according to our knowledge are the two best performing approaches in LRW. The first in proposed in \cite{chung2017lipsent}, and it deploys an encoder-decoder with temporal attention mechanism. Although the architecture is designed to address the problem of sentences-level recognition, it has also been tested on the LRW database, after fine-tuning on the LRW training set. The whole set of experimental results can be found in \cite{chung2017lipsent} and the results on LRW are repeated here in Table \ref{BL} (denoted by Watch-Attend-Spell). The second architecture is introduced by our team in \cite{Stafy2017} and its differences with the proposed one have been discussed above. The experimental results on LRW are given in Table \ref{BL} (denoted by ResNet-LSTM). Both experiment use the full set of words during training and evaluation (i.e. 500 words).
\begin{table}[!htbp]
\centering
\begin{tabular}{| c || c | c |} 
\hline
System & Top-1(\%) & Top-5(\%) \\ [0.5ex] 
\hline\hline 
Watch-Attend-Spell \cite{chung2017lipsent} & 23.80 & - \\ 
\hline 
ResNet-LSTM \cite{Stafy2017} & 17.03 & 3.72 \\
\hline
\end{tabular}
\vspace{1mm}
\caption{Baseline and state-of-the-art results on the full set (500 words).}
\label{BL}
\end{table}

\vspace{-0.7cm}     
\subsubsection{Experiments on the reduced set of words}
For the first experiment we use the proposed architecture without dropouts or batch normalization at the backend. The results are given in Table \ref{ID350} (denoted by N1) and the network attains $13.13$\% error rate on the reduced set. For the second experiment (denoted by N2) we add dropouts to the backend but again we do not apply batch normalization to the embedding layer. The error rate drops to $12.67$\%, showing the gains by applying dropouts at the backend. The next configuration uses both dropouts or batch normalization at the backend, and a single BiLSTM layer (denoted by N3). The network attains $12.59$\% error rate on the reduced set, showing that good results can be obtain even with a single BiLSTM layer. In the next configuration we experimented with extracting the embedding from the last output of the BiLSTM (as proposed in \cite{wand2016lipreading}), rather than with average pooling across all time steps. The network attains $12.15$\% error rate and it is denoted by N4. The next configuration is the proposed architecture without the use of word boundaries. The network (denotes by N5) attains $15.23$\%, showing that the network yields good results even without specifying the boundaries of the target words. Finally, the proposed architecture (denoted by N6) attains $11.29$\% error rate on the reduced set, which is clearly better that the other configurations examined. Moreover, by comparing N6 with N2 we notice the strength of batch normalization at the embedding layer. We should also mention that we experimented with the typical stacking approach of BiLSTM. In this case, the outputs of the first BiLSTM are concatenated and used as input to the second BiLSTM. The network failed to attain good results (error rates above $20$\%), despite our efforts to tune parameters such as learning rate and dropout probabilities.     

\begin{table}[!htbp]
\centering
\begin{tabular}{| P{0.4cm} |P{0.3cm}|P{0.45cm}|P{0.4cm}|P{0.4cm}|P{0.4cm} ||c | c |} 
\hline
Net & \#L & WB & DO & BN & EM & Top-1(\%) & Top-5(\%) \\ [0.5ex] 
\hline\hline
N1 & 2 & \checkmark &   &   & A & 13.13 & 2.26 \\ 
\hline 
N2  & 2 & \checkmark & \checkmark &   & A & 12.67 & 2.10 \\
\hline 
N3  & 1 & \checkmark & \checkmark & \checkmark & A &12.59 & 2.05 \\ 
\hline
N4 & 2 & \checkmark & \checkmark & \checkmark & L & 12.15 & 1.89 \\ 
\hline
N5  & 2 &  & \checkmark & \checkmark & A & 15.23 & 2.87 \\ 
\hline
N6  & 2 & \checkmark & \checkmark & \checkmark & A & 11.29 & 1.74 \\ 
\hline
\end{tabular}
\vspace{1mm}
\caption{Results on the reduced set (350 words) for various network configurations. Abbreviations: \#L: number of BLSTM layers, WB: use of word boundaries, DO: use of dropouts at the backend, BN: use of batch normalization at the embedding, EM: embedding extracted using average pooling (A) or from last time step (L).}
\label{ID350}
\end{table}

\vspace{-0.2cm}
\subsubsection{Experiments on the full set of words}  
\label{exp:FS}
The networks N5 and N6 are retrained from scratch and scored on the full set, and their performance is given in Table \ref{ID500}. Compared to the current state-of-the-art we observe an absolute improvement equal to $5.11$\% using about 2/3 of the parameters. The LSTM is indeed capable of learning how to use the word boundaries, without having to drop out out-of-boundaries frames or to apply frame masking. Even without word boundaries though, our new architecture yields $1.36$\% absolute improvement over \cite{Stafy2017}. Finally, our architecture halves the error rates attained by the baseline (attentional encoder-decoder, \cite{chung2017lipsent}).

\begin{table}[!htbp]
\centering
\begin{tabular}{| P{0.4cm} |P{0.3cm}|P{0.45cm}|P{0.4cm}|P{0.4cm}|P{0.4cm} ||c | c |}
\hline
Net & \#L & WB & DO & BN & EM & Top-1(\%) & Top-5(\%) \\ [0.5ex] 
\hline\hline 
N5  & 2 &  & \checkmark & \checkmark & A & 15.67 & 3.04 \\ 
\hline 
N6  & 2 & \checkmark & \checkmark & \checkmark & A & 11.92 & 1.94 \\
\hline
\end{tabular}
\vspace{-1mm}
\caption{Results on the full set (500 words) for various network configurations. Abbreviations same as in Table \ref{ID350}.}
\label{ID500}
\end{table}
\vspace{-0.5cm}
\subsection{Low-shot learning experiments}
\label{Sub:LS}
In this set of experiments we assess the capacity of the embeddings in generalizing to words unseen during training. To this end, we assume few instances for each of the 150 unseen words. These words are not included in the training set of the architecture (which is composed of 350 words) and they are merely deployed to estimate shallow word-dependent models on the embedding space. We design two experiments, namely closed-set identification and word matching. 
\subsubsection{PLDA modeling of embeddings}
We model the embeddings using PLDA, \cite{Ioffe2006}. We train a PLDA model with expectation-maximization on the set of 350 words, drawn from the test set of LRW (50 instances per word, i.e. 17500 training instances). PLDA is chosen due to its probabilistic nature, which enables us to form likelihood ratios, that are extensively used in biometric tasks, such as speaker and face verification, \cite{Kenny2013} \cite{PLDA2007} \cite{BayesianFace}. Its parameters are defined by ${\cal P} = (\mathbf{\mu},\mathbf{V},\mathbf{\Sigma})$, where $\mathbf{\mu}$ the mean value, $\mathbf{V}$ a matrix that models the word subspace and $\mathbf{\Sigma}$ a full symmetric positive definite matrix modelling the within-class variability. The PLDA generative model is the following
\begin{equation}
\mathbf{x}_{i} = \mu + \mathbf{V}\mathbf{y}_{c_{i}} + \mathbf{\epsilon}_{i},
\end{equation}
where $\mathbf{x}_{i}$ is an embedding in $\mathbb{R}^{d_{\mathbf{x}}}$ belonging to class $c_i$, $\mathbf{y}_{c} \sim N(\mathbf{0},\mathbf{I})$ is a random vector in $\mathbb{R}^{d_{\mathbf{y}}}$ shared by all instances of the same class, $\mathbf{\epsilon}_{i} \sim N(\mathbf{0},\mathbf{\Sigma})$ a random vector in $\mathbb{R}^{d_{\mathbf{x}}}$, and $d_{\mathbf{x}} \geq d_{\mathbf{y}}$. In the following experiments we use $d_{\mathbf{x}}=512$ and $d_{\mathbf{y}}=200$.
\subsubsection{Closed-set identification on unseen words}
We are interested in examining the performance of the embeddings in closed-set identification on the unseen set of words. To this end, the embeddings of the 150 unseen words are extracted. The overall number of available embeddings per word is 50 from the LRW validation set and another 50 from the test set. The validation set serves to estimate class-conditional density functions, based on the PLDA parameters, i.e. $p(\cdot|c,{\cal P}) = p(\cdot|\{\mathbf{x}_{i}\}_{c_{i}=c},{\cal P})$, where $c$ the class (i.e. word) label, $\{\mathbf{x}_{i}\}_{c_{i}=c}$ a set of instances from the validation set from class $c$.  

A class conditional density for each word $c$ given ${\cal P}$ is estimated using variable number of instances per word $N_{c}$ (from 1 to 16) drawn from the validation set of LRW. Subsequently, the models are evaluated on test embeddings (50 per word, from LRW test set) and the estimated word is derived using maximum likelihood. The Top-1 error rates for several number of training instances per word $N_{c}$ are given in Table \ref{LSL:ID} (denoted by ID-W350). For comparison, we include results where the embeddings are extracted from the network trained with the full set of 500 words (denoted by ID-W500), i.e. the one used in Sect. \ref{exp:FS}.  

\subsubsection{Word matching on unseen words}
For the final experiment, we evaluate log likelihood ratios (LLRs) between the hypotheses (a) the word instance $\mathbf{x}$ belongs to the same word-class $c$ with a collection of word instances $\{\mathbf{x}_{i}\}_{c_{i}=c}$, and (b) $\mathbf{x}$ and $\{\mathbf{x}_{i}\}_{c_{i}=c}$ belong to different classes. Contrary to closed-set identification, we assume that each instance may belong to an unknown set of classes. Moreover, since we are scoring pairs of word models and instances, more than one model per word can be created. We use again the validation set of LRW to create these models and the test set to create test instances. We measure the performance in terms of Equal Error Rate (EER), defined as the error rate attained when the LLRs are thresholded in such a way so that Missed Detection and False Alarm rates are equal. The results using variable number of training instances per model $N_c$ is given in Table \ref{LSL:ID} (denoted by EER-W350). For comparison, results with embeddings extracted from the same network trained on the full set of words are also given (denoted by EER-W500).
\begin{table}[!htbp]
\centering
\begin{tabular}{| c || c | c | c | c | c |}
\hline
 $N_{c}$ & 1 & 2 & 4 & 8 & 16  \\ [0.5ex] 
\hline\hline 
ID-W350 (\%) & $43.8$ & $32.5$ & $23.7$ & $19.2$ & $17.3$ \\ 
\hline
ID-W500 (\%) & $34.3$ & $23.0$ & $16.6$ & $13.1$ & $11.9$ \\ 
\hline\hline
EER-W350 (\%) & $6.11$ & $4.22$ & $3.31$ & $3.16$ & $3.03$ \\ 
\hline
EER-W500 (\%) & $4.52$ & $3.01$ & $2.49$ & $2.28$ & $2.21$ \\ 
\hline
\end{tabular}
\vspace{1mm}
\caption{Top-1 identification error and equal error rates on the unseen set of 150 words using PLDA for various training embeddings per word. W350 indicates that the network is trained on the reduced set while W500 on the full set.}
\label{LSL:ID}
\end{table}

Overall, the experiment on low-shot learning demonstrate the generalizability of the proposed architecture. Even with a modest number of training words (i.e. 350), the architecture succeeds in learning how to break down word instances into their ``visemic'' content and in extracting embeddings with good word discriminative properties.      
\section{Conclusion}
\label{sect:Conc}
In this paper, we proposed a deep learning architecture for lipreading that is capable of attaining performance beyond state-of-the-art in the challenging LRW database. The architecture combines spatiotemporal convolution, ResNets and LSTMs and an average pooling layer from which word embeddings are extracted. We explored several configurations of the LSTM-based back-end and we proposed an efficient method of using the word boundaries. We also attempted to address the problem of low-shot learning. To this end, we retrained the network on a subset of words (350 out of 500) and tested it on the remaining 150 words, using PLDA modelling. The experiments on low-shot learning show that good results can be attained even for words unseen during training.

For future work, we will train and test our architecture on the LipReading Sentences in-the-wild database (\cite{chung2017lipsent}) and we will experiment with word embeddings for large vocabulary visual speech recognition, using words as recognition units.  

\section{Acknowledgements}
This work has been funded by the European Commission program Horizon 2020, under grant agreement no. 706668 (Talking Heads).
\bibliographystyle{IEEEtran}
\bibliography{AV}

\end{document}